\documentclass[sigconf]{acmart}
\renewcommand\footnotetextcopyrightpermission[1]{} 
\AtBeginDocument{%
  }

\setcopyright{acmlicensed}
\copyrightyear{2018}
\acmYear{2018}
\acmDOI{XXXXXXX.XXXXXXX}
\acmConference[Conference acronym 'XX]{Make sure to enter the correct
  conference title from your rights confirmation email}{June 03--05,
  2018}{Woodstock, NY}
\acmISBN{978-1-4503-XXXX-X/2018/06}

\acmSubmissionID{5642}

\usepackage{graphicx}
\usepackage{enumitem}
\usepackage{comment}
\usepackage{booktabs}
\usepackage{multirow}
\usepackage{geometry}
\PassOptionsToPackage{table}{xcolor}
\usepackage{colortbl}
\usepackage{tikz}
\usepackage{pgfplots}
\pgfplotsset{compat=1.18}
\definecolor{color_zeroshot}{RGB}{220,210,195} 
\definecolor{color_sft}{RGB}{189,175,162}     
\definecolor{color_naive}{RGB}{154,175,169}   
\definecolor{color_agent}{RGB}{110,127,136}   
\usepackage{pifont}
\usepackage[ruled,vlined,linesnumbered]{algorithm2e}
\newcommand{\cmark}{\ding{51}}
\newcommand{\xmark}{\ding{55}}

\begin{document}

\title{Navigating the Emotion Tree: Hierarchical Hyperbolic RAG for Multimodal Emotion Recognition}


\author{Zeheng Wang}
\affiliation{%
  \institution{Great Bay University}
  \city{Guangdong, Dongguan}
  \country{China}}

\author{Bo Zhao}
\affiliation{%
  \institution{Great Bay University}
  \city{Guangdong, Dongguan}
  \country{China}}

\author{Yijie Zhu}
\affiliation{%
  \institution{Great Bay University}
  \city{Guangdong, Dongguan}
  \country{China}}

\author{Zhishu Liu}
\affiliation{%
  \institution{Great Bay University}
  \city{Guangdong, Dongguan}
  \country{China}}

\author{Hui Ma}
\affiliation{%
  \institution{Great Bay University}
  \city{Guangdong, Dongguan}
  \country{China}}

\author{Ruixin Zhang}
\affiliation{%
  \institution{Tencent Youtu Lab}
  \city{Shanghai}
  \country{China}}

\author{Shouhong Ding}
\affiliation{%
  \institution{Tencent Youtu Lab}
  \city{Shanghai}
  \country{China}}

\author{Qianyu Xie}
\affiliation{%
  \institution{CUHK, shenzhen}
  \city{Guangdong, Shenzhen}
  \country{China}}
  
\author{Zitong Yu}
\affiliation{%
  \institution{Great Bay University}
  \city{Guangdong, Dongguan}
  \country{China}}
\authornote{Corrsponding Author}
\renewcommand{\shortauthors}{Trovato et al.}

\begin{abstract}
Multimodal emotion recognition aims to integrate text, audio, and video sources to understand human affective states.
Although multimodal large language models excel at multimodal reasoning, they typically treat emotion categories as independent labels, ignoring the rich hierarchical taxonomy of human psychology.
Moreover, lacking external contextual knowledge makes them highly susceptible to over-interpreting noisy cues, further complicating fine-grained emotion classification.
To address these issues, we propose \textbf{HyperEmo-RAG}, a retrieval-augmented generation framework that leverages a structured emotional knowledge base. Our framework introduces two key innovations. 
\textbf{1) Hierarchical hyperbolic grounding.} Recognizing the inherent hierarchical tree structure of emotion taxonomies, we jointly embed hierarchical emotion labels and multimodal samples into a continuous hyperbolic space (Poincaré ball) and design a hierarchical beam-search deliberation process that progressively retrieves samples from coarse to fine-grained levels. 
\textbf{2) Structured evidence injection.} Based on the retrieved evidence, we construct an evidence graph and inject the structured knowledge as explicit cognitive context into the LLM through a Tree-Aware Attention mechanism and an EmotionGraphFormer, preserving the integrity of graph-structured information. 
Experiments on multiple datasets demonstrate that HyperEmo-RAG significantly outperforms existing methods.

\end{abstract}



\maketitle

\begin{figure}[t]
    \centering
    \includegraphics[width=\linewidth]{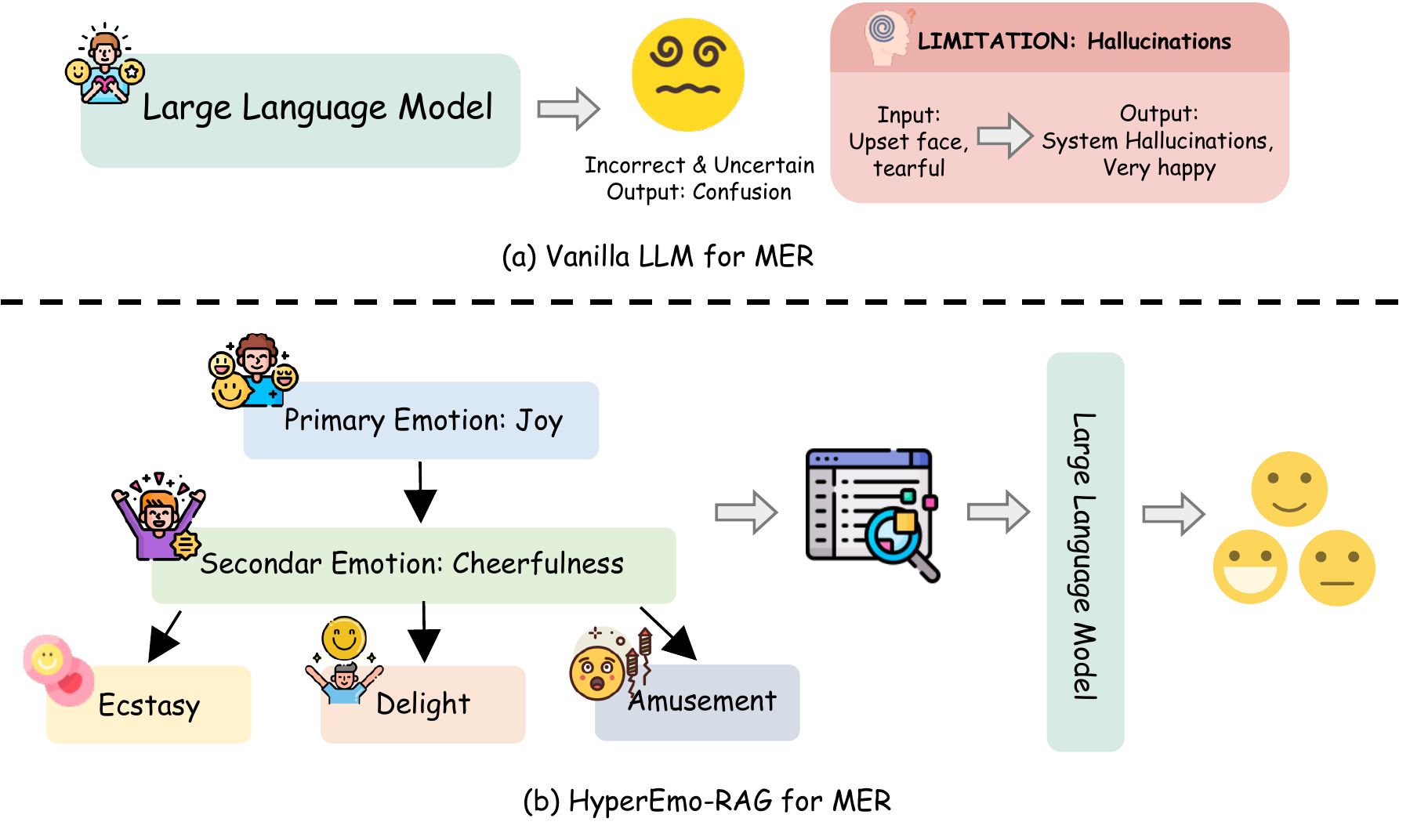} 
    \caption{Comparison of emotion recognition frameworks. \textbf{(a)} Traditional methods often treat emotion categories as flat labels and are susceptible to noise over-interpretation. \textbf{(b)} Our proposed HyperEmo-RAG jointly embeds a hierarchical emotion tree and multimodal samples into a hyperbolic space, leveraging retrieval-augmented generation to mitigate noise and enhance fine-grained recognition.}
    \label{fig:intro}
    \vspace{-2em}
\end{figure}
\section{Introduction}
Affective computing studies how machines perceive and interpret human emotions from heterogeneous signals such as language, voice, and facial behaviors \cite{cowie2001emotion, picard2001affective, poria2017review, soleymani2017survey, zadeh2018multimodal, zhu2025emosym, zhu2025uniemo, MIR-2025-09-543, liu2025llm,liu2026aullmstructuralreasoninglarge, wang2025tcgstriplanebasedcompression}. As intelligent systems are increasingly deployed in human-centered scenarios, multimodal emotion recognition (MER) has become an important research problem for applications such as education, counseling, and human-computer interaction. Compared with conventional perception tasks, MER is particularly challenging because emotional states are often conveyed through subtle, complementary, or even conflicting cues across text, audio, and video modalities.

Early progress in MER mainly relied on unimodal models \cite{li2022deep, zhao2022robust, wang2020region, schuller2018speech, el2011survey, devlin2019bert, lei2023instructerc, jia2022beyond} and multimodal fusion architectures \cite{zadeh2017tensor, liu2018efficient, tsai2020multimodal, rahman2020integrating, han2021improving, zhang2023learning, cheng2023semi}. Although these methods have improved feature extraction and cross-modal interaction, they remain largely constrained by feature-level alignment and fixed label supervision, which limits their ability to capture nuanced affective semantics. More recently, multimodal large language models have introduced a new paradigm for affective understanding by aligning heterogeneous modalities into a shared linguistic space \cite{lian2023affectgpt, cheng2024emotionllama}. Despite their promising reasoning ability, current LLM-based MER systems still tend to treat emotion categories as independent labels, overlooking the hierarchical emotions.

This limitation is especially evident in fine-grained emotion recognition, where emotions are not flat categories but hierarchically organized. Treating them as independent labels weakens a model's ability to distinguish closely related emotions. Meanwhile, without external evidence, large language models tend to over-rely on parametric memory and are prone to hallucinations. Retrieval-augmented generation (RAG) provides a solution by grounding predictions in external evidence \cite{abootorabi2025ask, yuan2025mrag, pipoli2025missrag, wen2025listen}. However, existing multimodal RAG methods usually adopt a single-round retrieve-then-generate paradigm and treat retrieved samples as an unordered memory pool. As a result, they remain inadequate for fine-grained emotion understanding.

To address these issues, we propose HyperEmo-RAG, a retrieval-augmented generation framework for multimodal emotion recognition grounded in a structured emotional knowledge base. Our method is built on two key ideas. First, since emotion taxonomies are inherently hierarchical, we introduce hierarchical hyperbolic grounding, which jointly embeds hierarchical emotion labels and multimodal samples into a continuous hyperbolic space (Poincar\'e ball). On top of this space, we design a hierarchical beam-search deliberation process that progressively retrieves evidence from coarse to fine-grained emotion levels, thereby alleviating retrieval error propagation and improving emotional disambiguation. Second, instead of treating retrieved instances as isolated examples, we organize them into a Deliberation Evidence Graph (DEG) that explicitly connects fine-grained emotion labels and retrieved multimodal evidence. We then introduce a Tree-Aware Attention mechanism and an EmotionGraphFormer to encode this graph and inject the resulting structured evidence into the LLM.

To further align multimodal representations with structured emotional concepts, HyperEmo-RAG is optimized with auxiliary objectives in hyperbolic space. Specifically, we use a tree-distance-aware multimodal contrastive loss to preserve semantic relations among emotion categories, and a path consistency loss to regularize the alignment between multimodal queries and selected fine-grained emotion prototypes. In this way, HyperEmo-RAG unifies hierarchical retrieval, structured evidence modeling, and knowledge-grounded generation within a coherent framework for MER. Our main contributions are summarized as follows:
\begin{itemize}[leftmargin=2em, topsep=4pt, itemsep=4pt, parsep=0pt]
    \item We propose \textbf{HyperEmo-RAG}, a retrieval-augmented generation framework for multimodal emotion recognition that grounds LLM reasoning with a structured emotional knowledge base.
    \item We construct a \textbf{Hyperbolic Emotion Tree}, which jointly embeds emotion taxonomies and multimodal samples in hyperbolic space, together with a hierarchical beam-search deliberation process for coarse-to-fine emotion retrieval.
    \item We develop a \textbf{structured evidence injection} mechanism based on a Deliberation Evidence Graph, Tree-Aware Attention, and EmotionGraphFormer, enabling retrieved emotional evidence to be incorporated into the LLM as structured context.
\end{itemize}
\section{Related Works}
{\flushleft \textbf{Multimodal Emotion Recognition.}} 
Multimodal Emotion Recognition (MER) aims to infer human affective states by jointly modeling textual, acoustic, and visual signals. Early studies mainly focused on unimodal modeling and multimodal fusion, improving performance through feature extraction, alignment, and interaction modeling across modalities \cite{zadeh2017tensor, tsai2020multimodal}. More recently, Multimodal Large Language Models (MLLMs) \cite{zhu2024minigpt, zhang2023videollama} have provided a unified semantic space for cross-modal understanding, opening a new direction for affective analysis. Instruction-tuned models such as AffectGPT \cite{lian2023affectgpt}, EmoLLM \cite{yang2024emollm}, and EmoLLaVA \cite{zheng2023emollava} further demonstrate the potential of large models in zero-shot and few-shot emotion-related tasks. Despite these advances, current MLLM-based MER systems still rely heavily on static parametric memory and typically treat emotion categories as independent labels. This makes them vulnerable to noisy or conflicting multimodal cues and limits their ability to capture the hierarchical structure and fine-grained boundaries of human emotions. These limitations suggest the need for MER frameworks that can incorporate structured external evidence beyond parametric memory.

{\flushleft \textbf{Multimodal Retrieval-Augmented Generation.}} 
Retrieval-Augmented Generation (RAG) alleviates the limitations of static model memory by grounding prediction and generation in external knowledge. Extending the text RAG paradigm, Multimodal RAG (M-RAG) retrieves cross-modal samples or knowledge to support multimodal understanding and generation \cite{chen2022murag, yasunaga2023racm3, hu2023reveal}. In affective computing, retrieved historical or external examples can provide useful reference anchors, improving the model's sensitivity to emotional context \cite{abootorabi2025ask, pipoli2025missrag, wang2026affectagent}. However, existing multimodal RAG methods are still not fully suitable for MER. Most follow a single-round retrieve-then-generate pipeline \cite{yuan2025mrag, jiang2023flare}, which is insufficient for emotion recognition, where disambiguation often requires progressive reasoning from coarse affective tendencies to fine-grained emotional states. In addition, retrieved samples are usually injected as an unordered set of examples, without explicitly modeling their structural relations to emotion concepts \cite{xia2024multimodal}. Therefore, current M-RAG methods still fall short of providing the hierarchical and structured grounding required for fine-grained multimodal emotion recognition. This gap motivates us to study a structured and hierarchical retrieval-augmented framework tailored to emotion understanding.
\begin{figure*}[t]
    \centering
    \includegraphics[width=\textwidth]{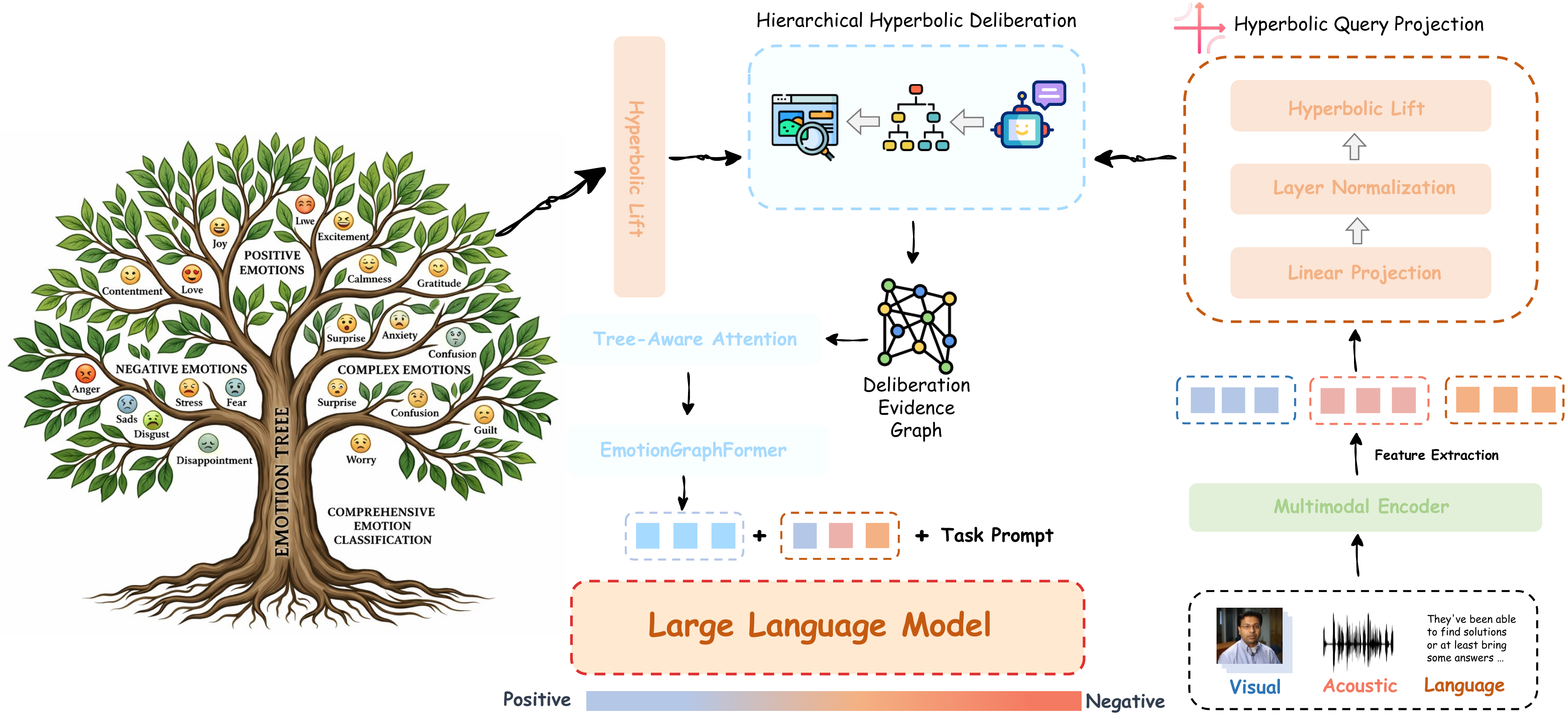}
    \caption{The overall framework of HyperEmo-RAG. Multimodal features (visual, acoustic, and language) are first extracted and projected into a hyperbolic space. Then, a hierarchical hyperbolic deliberation process is performed on the Emotion Tree to construct a Deliberation Evidence Graph. Finally, the EmotionGraphFormer distills the structured evidence, which is combined with task prompts and fed into a Large Language Model for comprehensive emotion classification.}
    \label{fig:framework}
    \vspace{-1em}
\end{figure*}
\section{Proposed Method}
Given a multimodal input sample $\mathcal{X}=\{x^a, x^v, x^t\}$, where $x^a$, $x^v$, and $x^t$ denote the audio, visual, and textual modalities, respectively, we first employ an upstream multimodal encoder to obtain modality-specific Euclidean features $\mathbf{h}_a$, $\mathbf{h}_v$, and $\mathbf{h}_t$. Since this front-end encoder is not the focus of this work, we keep it abstract and describe only the proposed \textbf{HyperEmo-RAG} modules built on top of these features. HyperEmo-RAG has three stages: (1) hyperbolic encoding of the emotion tree and query, (2) tree-based deliberative retrieval, and (3) graph-guided evidence injection. The overall framework is illustrated in Fig.~\ref{fig:framework}. Detailed procedure is summarized in Algorithm~\ref{alg:hyperemorag}.
\begin{algorithm}[th]
\small
\caption{Overview of HyperEmo-RAG}
\label{alg:hyperemorag}
\DontPrintSemicolon

\KwIn{Multimodal input $\mathcal{X}=\{x^a,x^v,x^t\}$, emotion tree $\mathcal{T}$, knowledge base $\mathcal{K}$}
\KwOut{Predicted emotion label $\hat{y}$}

Extract Euclidean features $\mathbf{h}_a,\mathbf{h}_v,\mathbf{h}_t$\;

Project them into hyperbolic space to obtain
$q_{\mathcal{X}}^a,q_{\mathcal{X}}^v,q_{\mathcal{X}}^t$\;

Fuse modality-specific queries into the sample query $q_{\mathcal{X}}$\;

Perform hierarchical deliberative retrieval on $\mathcal{T}$\;

Select a coarse-to-fine path $\pi^\ast$ and retrieve top-$k$ branch-consistent evidence from $\mathcal{K}$\;

Construct the Deliberation Evidence Graph $\mathcal{G}$ using
concept nodes on $\pi^\ast$ and retrieved evidence nodes\;

Encode $\mathcal{G}$ with Tree-Aware Attention\;

Use EmotionGraphFormer to distill $\mathcal{G}$ into $Q$ graph tokens $Z_g$\;

Combine multimodal prompt embeddings, graph tokens $Z_g$, and the task prompt to form the final LLM input\;

Feed the constructed input into the frozen LLM to obtain $\hat{y}$\;

\If{training}{
    Compute $L_{\mathrm{task}}$, $L_{\mathrm{cont}}$, and $L_{\mathrm{path}}$\;
    Optimize
    $
    L = L_{\mathrm{task}}
    + \lambda_{\mathrm{cont}}L_{\mathrm{cont}}
    + \lambda_{\mathrm{path}}L_{\mathrm{path}}
    $\;
}

\end{algorithm}
\vspace{-1em}

\subsection{Emotion Tree and Sample Encoding}
This stage contains two different types of representations: emotion-node representations and sample representations. The former describes the emotion hierarchy itself, while the latter describes the current multimodal input.

We first construct an emotion tree $\mathcal{T}$ based on the emotion hierarchy of the emotion wheel~\cite{lian2023affectgpt}. The tree $\mathcal{T}$ specifies how emotion labels are hierarchically related, while the prototype $\mathbf{p}_n$ provides a learnable anchor for node $n$ in the shared hyperbolic space. To make this hierarchy trainable, we assign each node $n$ a learnable prototype in hyperbolic space. Let $\mathbf{z}_n \in \mathbb{R}^{d_h}$ denote the Euclidean parameter of node $n$. Its hyperbolic prototype is obtained by:
\begin{equation}
\mathbf{p}_n = \mathrm{Exp}_{\mathbf{0}}(\mathbf{z}_n),
\end{equation}
where $\mathrm{Exp}_{\mathbf{0}}(\cdot)$ is the exponential map at the origin of the Poincar\'e ball. Here, $\mathbf{p}_n$ is the representation of the emotion concept associated with node $n$, rather than the representation of a particular input sample. In other words, the tree $\mathcal{T}$ specifies how emotion labels are hierarchically related, while the prototype $\mathbf{p}_n$ provides a learnable geometric anchor for node $n$ in the shared hyperbolic space.

Next, we encode the current multimodal input in the same space so that it can be compared with the emotion prototypes. Specifically, we apply three independent projection heads $f_a(\cdot)$, $f_v(\cdot)$, and $f_t(\cdot)$ to the audio, visual, and textual features, respectively. Each projection head is implemented as a linear layer followed by LayerNorm. The resulting modality-specific sample queries are:
\begin{equation}
\begin{aligned}
\mathbf{q}_a^{x} &= \mathrm{Exp}_{\mathbf{0}}(f_a(\mathbf{h}_a)), \\
\mathbf{q}_v^{x} &= \mathrm{Exp}_{\mathbf{0}}(f_v(\mathbf{h}_v)), \\
\mathbf{q}_t^{x} &= \mathrm{Exp}_{\mathbf{0}}(f_t(\mathbf{h}_t)),
\end{aligned}
\end{equation}
where $\mathbf{q}_a^{x}$, $\mathbf{q}_v^{x}$, and $\mathbf{q}_t^{x}$ denote the hyperbolic representations of the current sample in the audio, visual, and textual modalities, respectively.

To obtain a unified sample-level query, we further fuse these modality-specific queries in the tangent space. Let $\boldsymbol{\alpha}=[\alpha_a,\alpha_v,\alpha_t]=\mathrm{softmax}(\boldsymbol{\theta})$ denote the learnable modality weights, where $\alpha_a+\alpha_v+\alpha_t=1$. The fused sample query is defined as:
\begin{equation}
\mathbf{q}^{x}=
\mathrm{Exp}_{\mathbf{0}}
\left(
\alpha_a \mathrm{Log}_{\mathbf{0}}(\mathbf{q}_a^{x})
+\alpha_v \mathrm{Log}_{\mathbf{0}}(\mathbf{q}_v^{x})
+\alpha_t \mathrm{Log}_{\mathbf{0}}(\mathbf{q}_t^{x})
\right),
\end{equation}
where $\mathrm{Log}_{\mathbf{0}}(\cdot)$ is the logarithmic map at the origin. The fused query $\mathbf{q}^{x}$ is still a representation of the current input sample; it is not part of the emotion tree. Its role is to locate the sample on the emotion hierarchy by comparing it with the node prototypes $\mathbf{p}_n$. In contrast, the modality-specific queries $\mathbf{q}_a^{x}$, $\mathbf{q}_v^{x}$, and $\mathbf{q}_t^{x}$ are retained for modality-aware evidence retrieval in the next stage.

\subsection{Deliberative Retrieval}
We build an external emotion knowledge base offline. Each knowledge item stores audio, visual, and textual evidence embeddings together with its emotion label and associated caption metadata. Given the fused sample query $\mathbf{q}^{x}$ defined above, we do not retrieve evidence in a single flat step. Instead, we perform a search on the emotion tree.

The search starts from the upper levels of the hierarchy. We compare $\mathbf{q}^{x}$ with the prototypes of coarse emotion nodes and retain only a small number of plausible branches. The model then expands these candidate branches level by level, progressively moving from coarse affective tendencies to fine-grained emotional states. In this way, retrieval is formulated as a deliberation process: rather than making an early hard decision, the model preserves several candidate hypotheses and gradually refines them along the tree.

Within each retained branch, evidence is ranked with the modality-specific sample queries. Let $\mathbf{k}_i^a$, $\mathbf{k}_i^v$, and $\mathbf{k}_i^t$ denote the audio, visual, and textual embeddings of the $i$-th knowledge item in hyperbolic space. Its retrieval distance to the current sample is defined as
\begin{equation}
D_i=
\alpha_a d_{\mathbb{H}}(\mathbf{q}_a^{x},\mathbf{k}_i^a)
+\alpha_v d_{\mathbb{H}}(\mathbf{q}_v^{x},\mathbf{k}_i^v)
+\alpha_t d_{\mathbb{H}}(\mathbf{q}_t^{x},\mathbf{k}_i^t),
\end{equation}
where $d_{\mathbb{H}}(\cdot,\cdot)$ denotes the hyperbolic geodesic distance. This design preserves complementary evidence from different modalities instead of collapsing all signals into a single undifferentiated similarity score.

After the search reaches the fine-grained level, the deliberation module outputs three results for each sample: a coarse-to-fine emotion path, a final selected emotion prototype $\mathbf{p}^{*}$, and a small set of branch-consistent retrieved evidence items. These outputs are then used to construct structured evidence for the final prediction stage.

\subsection{Graph-Guided Evidence Injection}
The retrieved evidence is not directly flattened into tokens and appended to the LLM. The core motivation is that retrieved evidence should not be treated as an unordered memory pool; instead, the model should explicitly preserve structural information, such as which evidence item supports which emotion concept, and which retrieved items are mutually consistent. For this reason, we first organize the retrieval results into a \textbf{Deliberation Evidence Graph} before integrating them into the LLM. 

Given the coarse-to-fine emotion path and the retrieved evidence items from the previous stage, we construct a graph $\mathcal{G}=(\mathcal{V},\mathcal{E})$ with two types of nodes. The first type is concept nodes, initialized by the corresponding hyperbolic emotion prototypes. The second type is evidence nodes, initialized by the hyperbolic evidence embeddings retrieved from the external knowledge base. By introducing both types of nodes, the graph successfully bridges the symbolic emotion structure and the sample-level retrieved evidence. Regarding edge connections, we connect evidence nodes to the concept nodes they support, and we also connect evidence nodes that are close to each other in the hyperbolic space. Consequently, the DEG preserves both the global hierarchical structure of the emotion path and the local consistency among the retrieved evidence items.

Let $\mathbf{G}=\{\mathbf{g}_i\}_{i=1}^{N}$ denote the initial node features of the constructed graph, where $N$ is the total number of nodes. To enable the model to distinguish node identities and their exact positions within the hierarchy, we introduce a learnable node-type identifier $t_i$ and a hierarchy-level identifier $l_i$ for each node $i$. This initial augmentation process can be formulated as:
\begin{equation}
\mathbf{x}_i^{(0)}=\mathbf{g}_i+\mathbf{e}^{\text{type}}_{t_i}+\mathbf{e}^{\text{level}}_{l_i},
\end{equation}
where $\mathbf{e}^{\text{type}}_{t_i}$ and $\mathbf{e}^{\text{level}}_{l_i}$ denote the learnable type embedding and level embedding, respectively.

Subsequently, we apply \textbf{Tree-Aware Attention} to encode the graph. Since standard self-attention treats all nodes as fully connected, it easily blurs valid information. Therefore, we constrain message passing via graph connectivity to ensure that information is exchanged only along meaningful structural relations (i.e., the edge set $\mathcal{E}$). Specifically, we implement a multi-head self-attention (MHA) layer with an adjacency-constrained attention mask $\mathbf{M}$, followed by residual connections, LayerNorm, and a feed-forward network (FFN). The specific computations for this structure-aware encoding process are grouped as follows:
\begin{align}
M_{ij} &=
\begin{cases}
0, & i=j \ \text{or}\ (i,j)\in\mathcal{E},\\
-\infty, & \text{otherwise},
\end{cases} \\
\widehat{\mathbf{G}} &= \mathrm{MHA}(\mathbf{X}^{(0)},\mathbf{X}^{(0)},\mathbf{X}^{(0)};\mathbf{M}), \\
\widetilde{\mathbf{G}} &= \mathrm{FFN}\bigl(\mathrm{LayerNorm}(\mathbf{X}^{(0)}+\widehat{\mathbf{G}})\bigr).
\end{align}

Since the encoded graph $\widetilde{\mathbf{G}}$ is variable in size, feeding it directly into the frozen LLM is problematic. We thus introduce an \textbf{EmotionGraphFormer} to summarize the variable-sized graph into a fixed number of LLM-compatible tokens while distilling the most valuable graph-informed evidence. In practice, the model first maintains a set $\mathbf{Q}^{(0)}$ of $Q$ learnable query vectors. These queries selectively extract information from the graph nodes via cross-attention, interact with each other through self-attention, and are finally projected to the hidden dimension of the LLM to obtain the evidence tokens $\mathbf{Z}_g$. This series of distillation steps is formalized as:
\begin{align}
\mathbf{Q}^{(0)} &= \{\mathbf{u}_1,\mathbf{u}_2,\dots,\mathbf{u}_Q\}, \\
\mathbf{Q}^{(1)} &= \mathrm{CrossAttn}(\mathbf{Q}^{(0)},\widetilde{\mathbf{G}},\widetilde{\mathbf{G}}), \\
\mathbf{Q}^{(2)} &= \mathrm{SelfAttn}(\mathbf{Q}^{(1)});\mathbf{Z}_g =\mathbf{W}_{\mathrm{LLM}}\mathbf{Q}^{(2)}.
\end{align}

Finally, we inject the distilled graph-guided evidence into the input sequence of the LLM. Before this, the backbone network fuses multimodal features to generate a base prompt embedding $\mathbf{H}_0 = [\mathbf{H}_{\mathrm{fusion}}; \, \mathbf{H}_{\mathrm{text}}]$. We sequentially wrap this base prompt with multimodal-specific markers to form $\mathbf{H}_{\mathrm{mm}} = [E(\texttt{<Multimodal>}); \, \mathbf{H}_0; \, E(\texttt{</Multimodal>})]$, append the structure-aware evidence tokens $\mathbf{Z}_g$ generated by the EmotionGraphFormer to obtain $\mathbf{H}_{\mathrm{aug}} = [\mathbf{H}_{\mathrm{mm}}; \, \mathbf{Z}_g]$, and finally concatenate the task-specific prompt embedding $\mathbf{H}_{\mathrm{task}}$. The complete input sequence is thus constructed as:
\begin{equation}
\mathbf{H}_{\mathrm{in}} = [E(\texttt{<bos>}); \, \mathbf{H}_{\mathrm{aug}}; \, \mathbf{H}_{\mathrm{task}}].
\end{equation}

This complete sequence is ultimately fed into the frozen LLM for inference and prediction. During training, the embedding of the target label is appended after $\mathbf{H}_{\mathrm{in}}$ for supervised language-model learning. Through this decoupled design, the LLM can make accurate predictions based on both the backbone multimodal context and the externally retrieved graph-guided evidence, without the need to directly process complex underlying graph structures.

\subsection{Training Objective}
After graph-guided evidence injection, we jointly optimize the trainable modules around the frozen LLM. To achieve this, we design a comprehensive training objective comprising three distinct components: a task loss for the final prediction, a hyperbolic contrastive loss for structuring the emotion space, and a path consistency loss for aligning retrieval with the decision-making process. Specifically, the main task objective $\mathcal{L}_{\mathrm{task}}$ is formulated as the autoregressive negative log-likelihood given the final input embedding sequence $\mathbf{H}_{\mathrm{in}}$ and the target output sequence $\mathbf{y}=(y_1,\dots,y_T)$. To regularize the shared hyperbolic emotion space, we impose a multimodal contrastive loss $\mathcal{L}_{\mathrm{cont}}$ on the modality-specific sample queries. For each modality, samples with the same emotion label are pulled closer, while those with different labels are pushed apart, with the negative margin scaled by their semantic tree distance. Furthermore, a path consistency loss $\mathcal{L}_{\mathrm{path}}$ is introduced to measure the hyperbolic distance $d_{\mathbb{H}}(\cdot,\cdot)$ between the fused sample query $q^x$ and the prototype of the final selected emotion concept $p_{n^*}$. The specific formulations of these three individual objectives are grouped as follows:
\begin{align}
\mathcal{L}_{\mathrm{task}} &= -\sum_{t=1}^{T} \log p(y_t \mid y_{<t}, \mathbf{H}_{\mathrm{in}}), \\
\mathcal{L}_{\mathrm{cont}} &= \mathcal{L}_a+\mathcal{L}_v+\mathcal{L}_t, \\
\mathcal{L}_{\mathrm{path}} &= d_{\mathbb{H}}(q^x, p_{n^*}).
\end{align}

Ultimately, the overall training objective $\mathcal{L}$ is computed as the weighted sum of these three terms, balanced by coefficients $\lambda_{\mathrm{cont}}$ and $\lambda_{\mathrm{path}}$:
\begin{equation}
\mathcal{L} = \mathcal{L}_{\mathrm{task}} + \lambda_{\mathrm{cont}} \mathcal{L}_{\mathrm{cont}} + \lambda_{\mathrm{path}} \mathcal{L}_{\mathrm{path}}.
\end{equation}

In this way, the model is trained not only to make the correct final prediction, but also to organize emotion representations geometrically and maintain consistency between hyperbolic retrieval and downstream decision making.
\begin{table*}[htbp]
    \centering
    \caption{Experimental results for the Multimodal Sentiment Analysis (MSA) task on the MOSEI and SIMS-V2 datasets. The best-performing values are highlighted in \textbf{bold}. For the MOSEI dataset, results of models marked with an asterisk (*) are retrieved from the official benchmark repository, while the rest are extracted from their original publications. For the SIMS-V2 dataset, all baseline performances are sourced from \cite{liu2022make}.}
    \label{tab:main_results}
    \resizebox{\textwidth}{!}{
        \begin{tabular}{l ccccc ccccc}
            \toprule
            \multirow{2}{*}{Model} & \multicolumn{5}{c}{MOSEI} & \multicolumn{5}{c}{SIMS-V2} \\
            \cmidrule(lr){2-6} \cmidrule(lr){7-11}
            & Acc-2 & F1 & Acc-7 & MAE & Corr & Acc-2 & F1 & Acc2\_weak & MAE & Corr \\
            \midrule
            TFN*~\cite{zadeh2017tensor}            & 78.50 & 78.96 & 51.60 & 0.573 & 0.714 & 76.51 & 76.31 & 66.27 & 0.323 & 0.667 \\
            LMF*~\cite{liu2018efficient}           & 80.54 & 80.94 & 51.59 & 0.576 & 0.717 & 77.05 & 77.02 & 69.34 & 0.343 & 0.638 \\
            MulT*~\cite{tsai2019mult}              & 81.15 & 81.56 & 52.84 & 0.559 & 0.733 & 79.50 & 79.59 & 69.61 & 0.317 & 0.703 \\
            MAG-BERT*~\cite{rahman2020integrating} & 82.51 & 82.77 & 50.41 & 0.583 & 0.741 & 79.79 & 79.78 & 71.87 & 0.334 & 0.691 \\
            MISA~\cite{hazarika2020misa}           & 83.60 & 83.80 & 52.20 & 0.555 & 0.756 & 80.53 & 80.63 & 70.50 & 0.314 & 0.725 \\
            Self-MM*~\cite{yu2021learning}         & 82.81 & 82.53 & 53.46 & 0.530 & 0.765 & 79.01 & 78.89 & 71.87 & 0.335 & 0.640 \\
            MMIM~\cite{han2021improving}           & 82.24 & 82.66 & 54.24 & 0.526 & 0.772 & 80.95 & 80.97 & 72.28 & 0.316 & 0.707 \\
            AV-MC~\cite{liu2022make}               & -     & -     & -     & -     & -     & 82.50 & 82.55 & 74.54 & 0.297 & 0.732 \\
            CHFN~\cite{guo2022dynamically}         & 83.70 & 83.90 & 54.30 & 0.525 & 0.778 & -     & -     & -     & -     & -     \\
            UniMSE~\cite{hu2022unimse}             & 85.86 & 85.79 & 54.39 & 0.523 & 0.773 & -     & -     & -     & -     & -     \\
            $\text{UniSA}_{\text{GPT2}}$~\cite{li2023unisa} & 71.02 & -     & 41.36 & 0.838 & -     & -     & -     & -     & -     & -     \\
            $\text{UniSA}_{\text{T5}}$~\cite{li2023unisa}   & 84.22 & -     & 52.50 & 0.546 & -     & -     & -     & -     & -     & -     \\
            $\text{UniSA}_{\text{BART}}$~\cite{li2023unisa} & 84.93 & -     & 50.03 & 0.587 & -     & -     & -     & -     & -     & -     \\
            \midrule
            \textbf{HyperEmo-Qwen-1.8B}            & 86.43 & 86.19 & 53.87 & 0.533 & 0.756 & 82.89 & 82.36 & 75.90 & 0.293 & 0.702 \\
            \textbf{HyperEmo-LLaMA2-7B}            & 88.69 & 88.74 & \textbf{57.98} & \textbf{0.486} & \textbf{0.806} & 78.20 & 78.32 & 71.73 & 0.353 & 0.574 \\
            \textbf{HyperEmo-ChatGLM3-6B}          & \textbf{89.06} & \textbf{88.82} & 57.32 & 0.494 & 0.801 & \textbf{85.62} & \textbf{85.87} & \textbf{77.78} & \textbf{0.279} & \textbf{0.748} \\
            \bottomrule
        \end{tabular}
    }
    \vspace{-1em}
\end{table*}
\section{Experiment}
\subsection{Experimental Settings}
\begin{table}[htbp]
    \centering
    \caption{Experimental results of the Emotion Recognition in Conversations (ERC) task on the MELD and CHERMA datasets. The best-performing values are highlighted in \textbf{bold}. For the MELD dataset, results of models marked with an asterisk (*) are cited from \cite{hu2022mmdfn}, while the rest are extracted from their respective original publications. All baseline performances on the CHERMA dataset are sourced from \cite{sun-etal-2023-layer}.}
    \vspace{-1em}
    \label{tab:erc_results}
    \resizebox{\columnwidth}{!}{
        \begin{tabular}{l cc cc}
            \toprule
            \multirow{2}{*}{Model} & \multicolumn{2}{c}{MELD} & \multicolumn{2}{c}{CHERMA} \\
            \cmidrule(lr){2-3} \cmidrule(lr){4-5}
            & Acc & WF1 & Acc & WF1 \\
            \midrule
            TFN*~\cite{zadeh2017tensor}            & 60.77 & 57.74 & - & 68.37 \\
            LMF*~\cite{liu2018efficient}           & 61.15 & 58.30 & - & 68.23 \\
            EFT~\cite{tsai2019mult}                & - & - & - & 68.72 \\
            LFT~\cite{tsai2019mult}                & - & - & - & 69.05 \\
            MulT~\cite{tsai2019mult}               & - & - & - & 69.24 \\
            PMR~\cite{lv2021progressive}           & - & - & - & 69.53 \\
            LFMIM~\cite{sun2023layerwise}          & - & - & - & 70.54 \\
            MMGCN*~\cite{hu2021mmgcn}              & 60.42 & 58.31 & - & - \\
            MM-DFN*~\cite{hu2022mmdfn}             & 62.49 & 59.46 & - & - \\
            EmoCaps~\cite{li2022emocaps}           & - & 64.00 & - & - \\
            GA2MIF~\cite{li2023ga2mif}             & 61.65 & 58.94 & - & - \\
            UniMSE~\cite{hu2022unimse}             & 65.09 & 65.51 & - & - \\
            $\text{UniSA}_{\text{GPT2}}$~\cite{li2023unisa} & 48.12 & 31.26 & - & - \\
            $\text{UniSA}_{\text{T5}}$~\cite{li2023unisa}   & 64.52 & 62.17 & - & - \\
            $\text{UniSA}_{\text{BART}}$~\cite{li2023unisa} & 62.34 & 62.22 & - & - \\
            \midrule
            \textbf{HyperEmo-Qwen-1.8B}            & 64.49 & 62.72 & 72.83 & 72.33 \\
            \textbf{HyperEmo-LLaMA2-7B}            & 67.90 & 66.80 & 74.47 & 73.99 \\
            \textbf{HyperEmo-ChatGLM3-6B}          & \textbf{68.88} & \textbf{67.54} & \textbf{75.45} & \textbf{75.79} \\
            \bottomrule
        \end{tabular}
    }
    \vspace{-2em}
\end{table}

\noindent\textbf{Datasets.} To comprehensively evaluate HyperEmo-RAG, we conduct multimodal sentiment analysis (MSA) and emotion recognition in conversation (ERC) experiments on two English benchmarks and two Chinese benchmarks, namely MOSEI~\cite{zadeh-etal-2018-multimodal}, MELD~\cite{poria-etal-2019-meld}, SIMS-V2~\cite{liu2022make}, and CHERMA~\cite{sun-etal-2023-layer}.

\noindent\textbf{Implementation Details.} We evaluate HyperEmo-RAG on three large language backbones, namely Qwen-1.8B~\cite{bai2023qwen}, ChatGLM3-6B-base~\cite{glm2024chatglm}, and LLaMA2-7B~\cite{touvron2023llama2}. For clarity, we refer to the corresponding variants as HyperEmo-RAG-Qwen, HyperEmo-RAG-ChatGLM3, and HyperEmo-RAG-LLaMA2 when necessary. To ensure a fair comparison, we keep the overall HyperEmo-RAG pipeline unchanged across backbones, including the offline emotion knowledge base, hierarchical hyperbolic retrieval, graph-guided evidence encoding, and prompt-side evidence injection. Unless otherwise specified, the backbone LLM is kept frozen, and only the multimodal encoder, hyperbolic projection and deliberation modules, graph encoder, and prompt adaptation layers are optimized. Each setting is evaluated with five random seeds, and the reported results are averaged over the five runs. All experiments are conducted on a single V100 GPU. We optimize the trainable parameters using AdamW and adopt a cosine learning-rate schedule with a 10\% warmup ratio. Dataset-specific hyper-parameters, such as batch size, learning rate, and early stopping patience, are selected on the validation set for each dataset--backbone pair. To ensure a consistent target format across different backbone LLMs, we standardize the output representation for both regression and classification tasks. For sentiment regression, we explicitly prepend a ``+'' sign to non-negative labels during training so that the model learns a uniform signed numerical format. For emotion classification, each emotion category is mapped to a unique numerical index, and the label--index mapping is explicitly included in the task prompt. This design reduces output ambiguity and makes decoding more stable across different backbones and languages. To provide external affective knowledge without introducing evaluation leakage, we construct the offline emotion knowledge base from the training split of MER2025 using the corresponding MERCaptionPlus annotations, including the open-set emotion labels and reason descriptions. Each entry is further parsed into audio-oriented, video-oriented, and text-oriented descriptions and then encoded into a multimodal memory. During inference, the retrieved evidence is organized into graph-guided representations and injected into the language model through prompt-side conditioning. For the generic multimodal RAG comparison reported in Table~\ref{tab:generic_mrag}, all compared methods are instantiated on the representative Qwen-1.8B backbone under the same retrieval corpus, prompt budget, and training protocol.

\begin{sloppypar}
\noindent\textbf{Baselines.} We compare HyperEmo-RAG instantiated on Qwen-1.8B, LLaMA2-7B, and ChatGLM3-6B-base with a broad range of representative multimodal baselines, including TFN \cite{zadeh2017tensor}, LMF \cite{liu2018efficient}, MISA \cite{hazarika2020misa}, MAG-BERT \cite{rahman2020integrating}, Self-MM \cite{yu2021learning}, MMIM \cite{han2021improving}, CHFN \cite{guo2022dynamically}, UniMSE \cite{hu2022unimse}, UniSA-BART, UniSA-T5, and UniSA-GPT2 \cite{li2023unisa}, AV-MC \cite{liu2022make}, MMGCN \cite{hu2021mmgcn}, MM-DFN \cite{hu2022mmdfn}, EmoCaps \cite{li2022emocaps}, GA2MIF \cite{li2023ga2mif}, EFT, LFT, and MulT \cite{tsai2019mult}, PMR \cite{lv2021progressive}, and LFMIM \cite{sun2023layerwise}. We further compare HyperEmo-RAG with three representative generic multimodal RAG baselines, namely MuRAG \cite{chen2022murag}, RA-CM3 \cite{yasunaga2023racm3}, and REVEAL \cite{hu2023reveal}. These methods are adapted under the same multimodal backbone, retrieval corpus, prompt budget, and training protocol for controlled comparison.
\end{sloppypar}

\noindent\textbf{Metrics.} Because the benchmark datasets involve both regression and classification settings, we report task-specific evaluation metrics for each dataset. For MOSEI, we report MAE, Pearson correlation (Corr), seven-class accuracy (Acc-7), binary accuracy (Acc-2), and F1 score, where Acc-2 and F1 are computed under the non-negative versus negative criterion. For SIMS-V2, we report MAE, Corr, Acc2-weak, Acc-2, and F1 score, where Acc-2 and F1 are computed under the positive versus non-positive criterion, and Acc2-weak is used to further evaluate weak-emotion instances within the interval [-0.4, 0.4]. For MELD and CHERMA, we report classification accuracy (Acc) and weighted F1 (WF1). For the generic multimodal RAG comparison in Table~\ref{tab:generic_mrag}, we retain the same task-specific metrics, namely MAE, Corr, Acc-7, Acc-2, and F1 for MOSEI; MAE, Corr, Acc2$_w$, Acc-2, and F1 for SIMS-V2; and Acc and WF1 for MELD and CHERMA.

\begin{table*}[ht]
    \centering
    \scriptsize
    \setlength{\tabcolsep}{3.0pt}
    \renewcommand{\arraystretch}{1.10}
    \caption{Comparison with representative generic multimodal RAG baselines on the Qwen-1.8B backbone. All compared methods are adapted under the same retrieval corpus, prompt budget, and training protocol for controlled comparison. Lower is better for MAE, and higher is better for all other metrics. Best results are highlighted in bold.}
    \vspace{-1em}
    \label{tab:generic_mrag}
    \resizebox{\textwidth}{!}{
    \begin{tabular}{@{}c ccccc ccccc cc cc@{}}
        \toprule
        \multicolumn{1}{c}{Method} & \multicolumn{5}{c}{MOSEI} & \multicolumn{5}{c}{SIMS-V2} & \multicolumn{2}{c}{MELD} & \multicolumn{2}{c}{CHERMA} \\
        \cmidrule(lr){2-6}\cmidrule(lr){7-11}\cmidrule(lr){12-13}\cmidrule(lr){14-15}
        & MAE$\downarrow$ & Corr$\uparrow$ & Acc-7$\uparrow$ & Acc-2$\uparrow$ & F1$\uparrow$
        & MAE$\downarrow$ & Corr$\uparrow$ & Acc2$_w$$\uparrow$ & Acc-2$\uparrow$ & F1$\uparrow$
        & Acc$\uparrow$ & WF1$\uparrow$ & Acc$\uparrow$ & WF1$\uparrow$ \\
        \midrule
        MuRAG~\cite{chen2022murag}
        & 0.579 & 0.815 & 53.14 & 87.84 & 87.49
        & 0.399 & 0.476 & 76.71 & 80.18 & 79.82
        & 66.96 & 66.61 & 75.04 & 74.88 \\

        RA-CM3~\cite{yasunaga2023racm3}
        & 0.555 & 0.834 & 54.06 & 88.31 & 88.02
        & 0.376 & 0.507 & 77.19 & 80.91 & 80.49
        & 67.41 & 67.08 & 75.47 & 75.22 \\

        REVEAL~\cite{hu2023reveal}
        & 0.563 & 0.828 & 53.79 & 88.11 & 87.80
        & 0.383 & 0.496 & 77.05 & 80.63 & 80.22
        & 67.28 & 66.95 & 75.31 & 75.07 \\
        \midrule
        HyperEmo-RAG (Ours)
        & \textbf{0.537} & \textbf{0.845} & \textbf{54.72} & \textbf{88.79} & \textbf{88.57}
        & \textbf{0.362} & \textbf{0.521} & \textbf{77.78} & \textbf{81.42} & \textbf{81.05}
        & \textbf{67.88} & \textbf{67.53} & \textbf{76.14} & \textbf{75.71} \\
        \bottomrule
    \end{tabular}}
    \vspace{-1em}
\end{table*}
\subsection{Main Results}
As shown in Tables~\ref{tab:main_results} and~\ref{tab:erc_results}, HyperEmo-RAG delivers competitive and generally stronger results than prior multimodal baselines on both sentiment analysis benchmarks, although the gains are not identical across backbones or metrics. On MOSEI, HyperEmo-ChatGLM3-6B achieves the best Acc-2/F1 of 89.06/88.82, outperforming the strongest prior classification baseline UniMSE by 3.20/3.03 points, while HyperEmo-LLaMA2-7B attains the best Acc-7, MAE, and Corr of 57.98/0.486/0.806, improving over the strongest reported baselines by 3.59 points in Acc-7, 0.037 in MAE, and 0.028 in Corr. This pattern suggests that the proposed framework is beneficial to both binary decision boundaries and finer-grained sentiment estimation, but the specific strength depends on the backbone. A similar trend is observed on SIMS-v2, where HyperEmo-ChatGLM3-6B performs best on all reported metrics, surpassing AV-MC by 3.12 points in Acc-2, 3.32 points in F1, 3.24 points in Acc2-weak, while also reducing MAE from 0.297 to 0.279 and improving Corr from 0.732 to 0.748. Notably, the gain on weak-emotion instances indicates that the retrieved emotional evidence is particularly useful when the sentiment signal is subtle, although the weaker LLaMA2 results on SIMS-v2 also show that the benefit is not completely uniform and may depend on backbone--dataset alignment.

The results on MELD and CHERMA further suggest that the advantage of HyperEmo-RAG is not limited to utterance-level sentiment prediction, but can also transfer to conversational emotion recognition. On MELD, HyperEmo-ChatGLM3-6B achieves the best Acc/WF1 of 68.88/67.54, exceeding UniMSE by 3.79/2.03 points, and HyperEmo-LLaMA2-7B also outperforms the strongest previous baseline on both metrics. On CHERMA, HyperEmo-ChatGLM3-6B obtains the best Acc/WF1 of 75.45/75.79, and its WF1 exceeds the strongest reported baseline LFMIM by 5.25 points. These improvements suggest that hyperbolic retrieval and graph-guided evidence injection provide complementary emotional context that is useful for disambiguating conversational affect. At the same time, the results should be interpreted with some caution: HyperEmo-Qwen-1.8B is slightly weaker than UniMSE on MELD, and prior work on CHERMA mainly reports WF1 rather than a full metric suite, which limits the breadth of direct comparison. Overall, the evidence supports a moderate conclusion that HyperEmo-RAG improves robustness across diverse multimodal affective tasks, while the extent of improvement still depends on the underlying backbone and benchmark characteristics.

\vspace{-1em}

\subsection{Comparison with RAG Methods}
Table~\ref{tab:generic_mrag} compares HyperEmo-RAG with three representative generic multimodal RAG baselines on the Qwen-1.8B backbone. Overall, HyperEmo-RAG achieves the best results across all four datasets. The generic multimodal RAG baselines already provide competitive performance, which suggests that external multimodal knowledge is useful for emotion understanding. Among them, RA-CM3 is generally the strongest baseline, while MuRAG and REVEAL also yield stable improvements over non-retrieval settings.

Compared with these generic RAG methods, HyperEmo-RAG shows consistent gains on both regression-oriented and classification-oriented benchmarks. A likely reason is that generic multimodal RAG mainly retrieves related multimodal evidence, but does not explicitly organize this evidence according to hierarchical emotional relations. In contrast, HyperEmo-RAG introduces emotion-oriented knowledge grounding and structured evidence integration, which helps the model use retrieved information in a way that is more aligned with fine-grained emotion understanding. The improvements are moderate but consistent, indicating that the proposed retrieval design is beneficial beyond generic multimodal retrieval alone.

\begin{table}[t]
    \centering
    \caption{Ablation on evidence channels with HyperEmo-ChatGLM3-6B. “Captions” denotes retrieved caption evidence, “Graph Tokens” denotes graph-token injection, and “Random Evid.” replaces both channels with random evidence of the same size.}
    \vspace{-1em}
    \label{tab:ablation_channels}
    \resizebox{\columnwidth}{!}{
        \begin{tabular}{ccc c ccc c}
            \toprule
            \multirow{2}{*}{Captions} & \multirow{2}{*}{Graph Tokens} & \multirow{2}{*}{Random Evid.} & MOSEI & \multicolumn{2}{c}{SIMS-V2} & MELD & CHERMA \\
            \cmidrule(lr){4-4} \cmidrule(lr){5-6} \cmidrule(lr){7-7} \cmidrule(lr){8-8}
            & & & F1 & F1 & Acc2\_w & WF1 & WF1 \\
            \midrule
            \xmark & \xmark & \xmark & 86.72 & 83.18 & 74.26 & 64.73 & 73.96 \\
            \xmark & \xmark & \cmark & 86.48 & 82.95 & 73.92 & 64.31 & 73.65 \\
            \cmark & \xmark & \xmark & 87.96 & 84.94 & 76.21 & 66.11 & 74.86 \\
            \xmark & \cmark & \xmark & 87.68 & 84.66 & 75.74 & 65.87 & 74.72 \\
            \midrule
            \cmark & \cmark & \xmark & \textbf{88.82} & \textbf{85.87} & \textbf{77.78} & \textbf{67.54} & \textbf{75.79} \\
            \bottomrule
        \end{tabular}
    }
\end{table}
\subsection{Ablation Study}
All ablation experiments are conducted with HyperEmo-ChatGLM3-6B under the same training and evaluation settings as the main experiments, and each variant changes only the target component to enable a controlled comparison. Tables~\ref{tab:ablation_channels}, ~\ref{tab:ablation_arch}, and ~\ref{tab:ablation_retrieval} examine the contribution of the evidence channels, graph-related architectural components and auxiliary objectives, and the retrieval strategy, respectively. In Table~\ref{tab:ablation_channels}, "Backbone only" removes both retrieved captions and graph-token injection, "Random evidence" replaces the retrieved evidence with randomly sampled evidence of comparable size, and the remaining variants retain either the caption channel or the graph-token channel alone. The results show that both evidence channels improve over the backbone-only setting, while the full model achieves the best performance on all benchmarks. In particular, the random-evidence variant does not improve over the backbone-only baseline and is slightly worse on several datasets, which suggests that the gains are not simply caused by a longer input sequence. At the same time, using only retrieved captions or only graph tokens already brings measurable improvements, whereas combining the two yields further gains, indicating that the textual evidence and the structured graph evidence provide complementary information.
\begin{figure*}[!t]
    \centering
    \includegraphics[width=\textwidth]{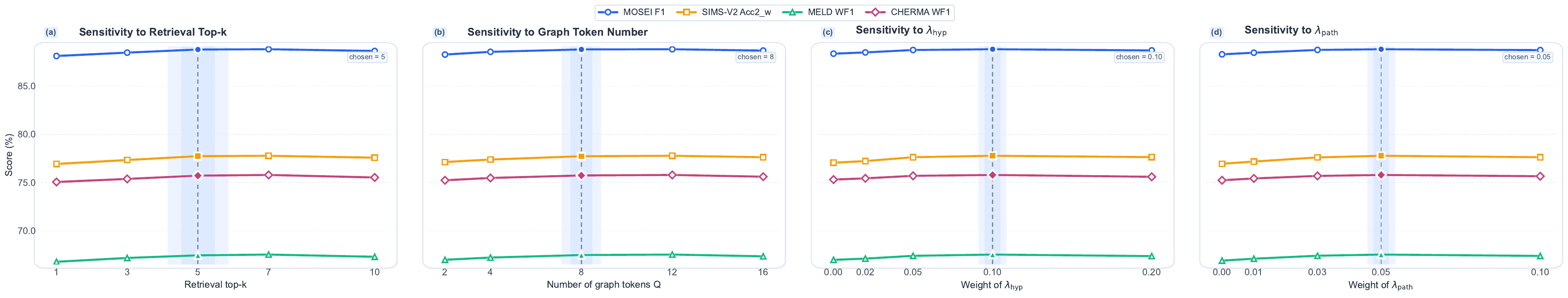}
    \caption{Parameter sensitivity analysis of HyperEmo-RAG with respect to the retrieval top-k, the number of graph tokens, the weight of $\lambda_{\mathrm{hyp}}$, and the weight of $\lambda_{\mathrm{path}}$. All other settings are kept unchanged, and the model generally achieves the best performance within a moderate parameter range.}
    \label{fig:param_sensitivity}
    \vspace{-1em}
\end{figure*}
\begin{table}[t]
    \centering
    \caption{Ablation on architectural components and auxiliary losses. We disable one component or one training objective at a time while keeping the remaining pipeline unchanged.}
    \vspace{-1em}
    \label{tab:ablation_arch}
    \resizebox{\columnwidth}{!}{
        \begin{tabular}{cccc c ccc c}
            \toprule
            \multirow{2}{*}{Tree Attn.} & \multirow{2}{*}{GraphFormer} & \multirow{2}{*}{$\mathcal{L}_{\text{hyp}}$} & \multirow{2}{*}{$\mathcal{L}_{\text{path}}$} & MOSEI & \multicolumn{2}{c}{SIMS-V2} & MELD & CHERMA \\
            \cmidrule(lr){5-5} \cmidrule(lr){6-7} \cmidrule(lr){8-8} \cmidrule(lr){9-9}
            & & & & F1 & F1 & Acc2\_w & WF1 & WF1 \\
            \midrule
            \xmark & \cmark & \cmark & \cmark & 88.12 & 85.06 & 76.68 & 66.82 & 75.12 \\
            \cmark & \xmark & \cmark & \cmark & 88.05 & 84.89 & 76.53 & 66.65 & 74.98 \\
            \cmark & \cmark & \xmark & \cmark & 88.36 & 85.31 & 77.06 & 66.99 & 75.31 \\
            \cmark & \cmark & \cmark & \xmark & 88.29 & 85.22 & 76.95 & 66.91 & 75.24 \\
            \cmark & \cmark & \xmark & \xmark & 87.91 & 84.76 & 76.39 & 66.34 & 74.73 \\
            \midrule
            \cmark & \cmark & \cmark & \cmark & \textbf{88.82} & \textbf{85.87} & \textbf{77.78} & \textbf{67.54} & \textbf{75.79} \\
            \bottomrule
        \end{tabular}
    }
    \vspace{-1em}
\end{table}
Table~\ref{tab:ablation_arch} further studies whether the performance gain depends on the way graph evidence is encoded and optimized. Starting from the full model, we remove Tree-Aware Attention, EmotionGraphFormer, $\mathcal{L}_{\mathrm{hyp}}$, or $\mathcal{L}_{\mathrm{path}}$ one at a time, and we also test a variant without both auxiliary losses. Removing Tree-Aware Attention or EmotionGraphFormer leads to a consistent drop across all datasets, which indicates that the graph branch is useful not only because additional evidence is introduced, but also because the structural relations inside the deliberation evidence graph need to be encoded in a suitable way before being injected into the language model. Removing either auxiliary loss also causes a smaller but still consistent decline, and removing both losses leads to a larger degradation. This pattern suggests that the main improvement comes from the evidence retrieval and graph-guided injection pipeline itself, while the auxiliary losses provide additional regularization that helps stabilize and refine representation learning.

\begin{table}[t]
    \centering
    \caption{Ablation on the retrieval mechanism. "Euclid." denotes Euclidean retrieval, "Hyperb." denotes hyperbolic retrieval, and "Hier. Delib." denotes hierarchical deliberation.}
    \vspace{-1em}
    \label{tab:ablation_retrieval}
    \resizebox{\columnwidth}{!}{
        \begin{tabular}{ccc c ccc c}
            \toprule
            \multirow{2}{*}{Euclid.} & \multirow{2}{*}{Hyperb.} & \multirow{2}{*}{Hier. Delib.} & MOSEI & \multicolumn{2}{c}{SIMS-V2} & MELD & CHERMA \\
            \cmidrule(lr){4-4} \cmidrule(lr){5-6} \cmidrule(lr){7-7} \cmidrule(lr){8-8}
            & & & F1 & F1 & Acc2\_w & WF1 & WF1 \\
            \midrule
            \cmark & \xmark & \xmark & 88.35 & 85.12 & 76.85 & 66.94 & 75.11 \\
            \xmark & \cmark & \xmark & 88.58 & 85.45 & 77.29 & 67.21 & 75.43 \\
            \midrule
            \xmark & \cmark & \cmark & \textbf{88.82} & \textbf{85.87} & \textbf{77.78} & \textbf{67.54} & \textbf{75.79} \\
            \bottomrule
        \end{tabular}
    }
    \vspace{-2.5em}
\end{table}
Finally, Table~\ref{tab:ablation_retrieval} evaluates the retrieval mechanism by comparing Euclidean retrieval, hyperbolic retrieval without hierarchical deliberation, and the full hierarchical hyperbolic retrieval. Under the same backbone and optimization setting, the Euclidean variant gives the weakest results, replacing it with hyperbolic retrieval yields consistent improvements, and adding hierarchical deliberation further improves performance on all reported metrics. For example, on SIMS-V2, Acc2$_w$ increases from 76.85 to 77.29 and then to 77.78, and on CHERMA, WF1 increases from 75.11 to 75.43 and then to 75.79. These results suggest that the retrieval design contributes at two levels: the hyperbolic space is better aligned with hierarchical emotional relations, and the multi-step deliberation process further improves the quality of the retrieved evidence. More specifically, hyperbolic retrieval improves the geometric alignment between multimodal samples and hierarchically related emotion concepts, while hierarchical deliberation further enhances evidence selection by progressively filtering candidate branches from coarse to fine levels. At the same time, the gains are moderate rather than extreme, so a more careful conclusion is that the hierarchical hyperbolic retrieval strengthens the full system in a complementary way instead of acting as the only source of performance improvement.
\vspace{-1em}

\subsection{Sensitivity Analysis}
We further analyze the sensitivity of HyperEmo-RAG to several key hyperparameters, including the retrieval top-$k$, the number of graph tokens $Q$, the weight of the hyperbolic contrastive loss $\lambda_{\mathrm{hyp}}$, and the weight of the path consistency loss $\lambda_{\mathrm{path}}$. As shown in Fig.~\ref{fig:param_sensitivity}, in each experiment we vary only one hyperparameter while keeping the backbone, training protocol, data split, and all other settings fixed. Following the main experiments, we report representative metrics on four benchmarks, namely MOSEI F1, SIMS-V2 Acc2$_w$, MELD WF1, and CHERMA WF1.

Overall, the model remains stable within a moderate parameter range. As the retrieval top-$k$ increases from a small value, performance consistently improves on all datasets, indicating that too few retrieved evidence items are insufficient for robust prediction. However, when too many items are retained, the gains gradually saturate and then slightly decline, suggesting that excessive retrieval may introduce redundant or weakly related information. A similar trend is observed for the number of graph tokens: increasing $Q$ from a very small value improves performance by preserving more structural information from the deliberation evidence graph, whereas overly large $Q$ brings limited additional benefit and may consume more budget in the language model.

The auxiliary loss weights show a similar moderate-range preference. As $\lambda_{\mathrm{hyp}}$ increases from 0 to a moderate value, performance improves steadily, suggesting that the hyperbolic alignment objective helps organize multimodal representations according to hierarchical emotional relations. However, an overly large weight leads to a slight drop, indicating that excessive regularization may interfere with the main task optimization. The sensitivity curve of $\lambda_{\mathrm{path}}$ follows the same pattern: a moderate path consistency weight provides effective structural guidance, whereas too small a value is insufficient and too large a value slightly restricts task-specific adaptation.

These hyperparameters also play different roles in the overall framework. Specifically, retrieval top-$k$ mainly controls the coverage--noise trade-off of external evidence, $Q$ determines the capacity of graph-structured evidence distilled into the LLM, while $\lambda_{\mathrm{hyp}}$ and $\lambda_{\mathrm{path}}$ primarily regulate geometric structure learning and retrieval--decision consistency, respectively. The overall trend suggests that HyperEmo-RAG performs best when evidence coverage and structural regularization remain well balanced, rather than being pushed to extremes.

Based on these observations, we adopt $k=5$, $Q=8$, $\lambda_{\mathrm{hyp}}=0.10$, and $\lambda_{\mathrm{path}}=0.05$ in the final model. Therefore, HyperEmo-RAG is not overly sensitive to exact hyperparameter choices, although keeping them within a moderate and reasonable range remains important for achieving the best performance.

\section{Conclusion}
We proposed HyperEmo-RAG, a retrieval-augmented framework for multimodal emotion recognition that explicitly models the hierarchical structure of emotions. By combining hierarchical hyperbolic retrieval with graph-guided evidence injection, the proposed method enables coarse-to-fine emotional deliberation and more effective use of external multimodal evidence for fine-grained emotion understanding. Extensive experiments on multiple English and Chinese benchmarks demonstrate that HyperEmo-RAG consistently improves performance over strong multimodal baselines and representative generic multimodal RAG methods. These results suggest that structured hierarchical grounding is a promising direction for building more robust and interpretable multimodal emotion recognition systems.

\bibliographystyle{ACM-Reference-Format}
\bibliography{sample-base}

\end{document}